\documentclass[10pt,twocolumn,letterpaper]{article}

\usepackage{cvpr}              

\usepackage{graphicx}
\usepackage{amsmath}
\usepackage{amssymb}
\usepackage{booktabs}
\usepackage{todonotes}
\usepackage{enumitem}
\usepackage{svg}
\usepackage{tabularray}

\usepackage[pagebackref,breaklinks,colorlinks]{hyperref}

\usepackage[capitalize]{cleveref}
\crefname{section}{Sec.}{Secs.}
\Crefname{section}{Section}{Sections}
\Crefname{table}{Table}{Tables}
\crefname{table}{Tab.}{Tabs.}

\def\cvprPaperID{*****} 
\def\confName{Conference}
\def\confYear{2023}

\begin{document}

\title{Deep Learning Approach for Knee Point Detection on Noisy Data}

\author{Ting Yan Fok, Nong Ye\\
Arizona State University\\
School of Computing and Augmented Intelligence\\
{\tt\small \{tfok2, nongye\}@asu.edu}
}
\maketitle

\begin{abstract}
    A knee point on a curve is the one where the curve levels off after an increase. In a computer system, it marks the point at which the system's performance is no longer improving significantly despite adding extra resources. Thus a knee point often represents an optimal point for decision. However, identifying knee points in noisy data is a challenging task.     
    All previous works defined knee points based on the data in the original scale. However, in this work, we define knee points based on normalized data and provide a mathematical definition of curvature for normalized discrete data points, based on the mathematical definition of curvature for continuous functions. The impact of normalization exerted on curvature and the location of knee points are also discussed. Nevertheless, assessing the effectiveness of methods is difficult in the absence of ground truth data and benchmark datasets, which makes comparing existing methods challenging. In view of this, we create synthetic data that simulate real-world scenarios. We achieve this by selecting a set of functions that possess the required characteristics in this research and then introducing noise that satisfies the underlying distribution. In addition, we present a deep-learning approach and employ a Convolutional Neural Network (CNN) with a U-Net-like architecture, to accurately detect the knee point(s) of the underlying true distribution. The proposed model is evaluated against state-of-the-art methods. Experiments show that our network outperforms existing methods in all synthetic datasets, regardless of whether the samples have single or multiple knee points. In fact, our model achieves the best $F_{1}$ scores among all existing methods in all the test sets.       
\end{abstract}

\section{Introduction}
\label{sec:intro}

Researchers in various fields frequently encounter the task of identifying knees/elbows. In this context, ”knees” are points where the concavity of a curve is negative (concave downward), while ”elbows” are points where the concavity is positive (concave upward). Generally speaking, a knee point represents an advantageous operation point that optimizes the balance between system performance and operational costs. Therefore, a reliable and precise knee/elbow point detection method is desired as selecting the ”right” operating point can lead to efficient utilization of system resources, which in turn results in cost savings and performance benefits. In the field of system
behavior, a knee point is a point at which the cost of altering system parameters is no longer justified by the expected performance benefit. This concept can be observed in the Network Congestion Control problem, where an ideal sending rate is desired to ensure fair traffic share and prevent  congestion. If the curve of packet delay increases significantly and then levels off, indicating there is network congestion, the protocol should halt increasing the sending rate. In the application of lithium-ion batteries,
the knee point on a Capacity Fade Curve hints the beginning of lithium-ion cell degradation and the battery is approaching its End-of-Life \cite{li-ion-cell_9, li-ion-cell_10, li-ion-cell_13, li-ion-cell_15}. In the application of BotNet Detection, a knee point can help identify potential controllers used by a master host to relay to bots. When it comes to clustering applications, the Elbow method is one of the most popular approaches for determining the ideal number of clusters. The elbow point on the plot of an evaluation criteria curve, such as the within-cluster sum of squares as a function of the number of clusters, represents the ideal number of clusters. Choosing the appropriate number of clusters can help in preventing over-fitting and ensuring precise outcomes.

\begin{figure*}
  \hspace*{\fill}%
  \begin{subfigure}{0.25\linewidth}
      \includegraphics[width=400pt]{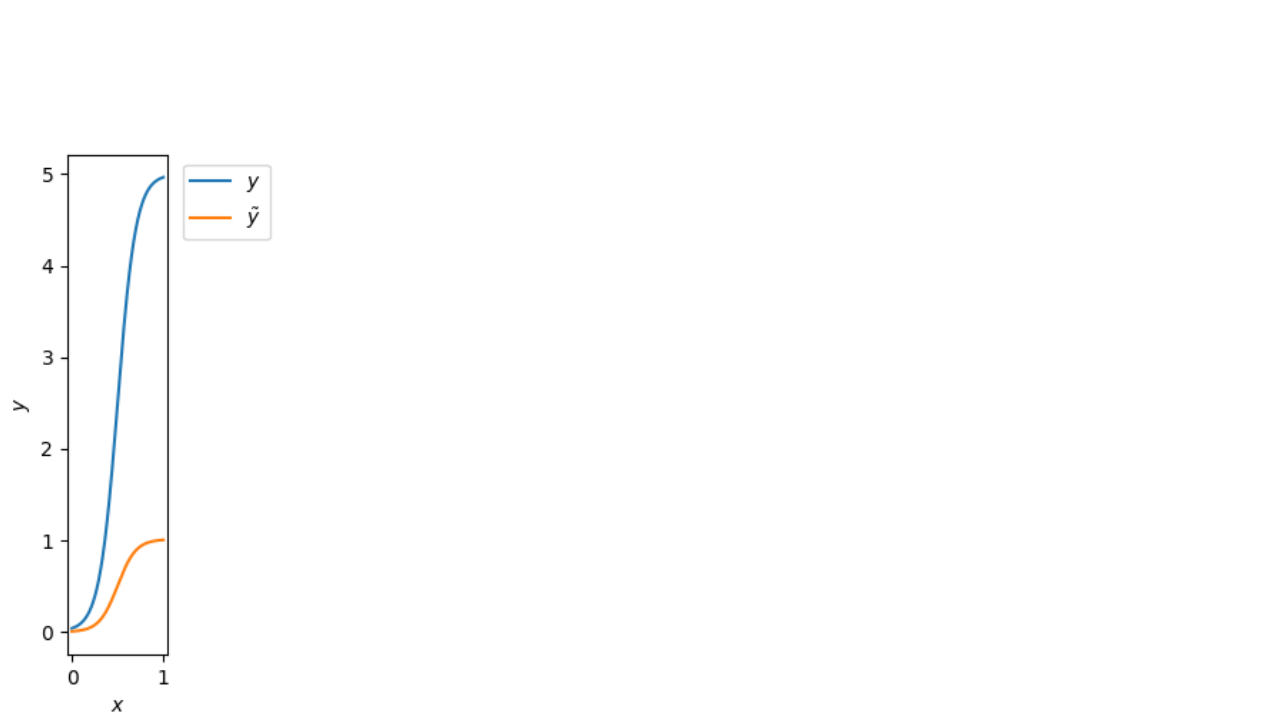}
   \subcaption{}
    \label{fig:sig_org}
  \end{subfigure}
  \begin{subfigure}{0.74\linewidth}
    \includegraphics[width=\textwidth]{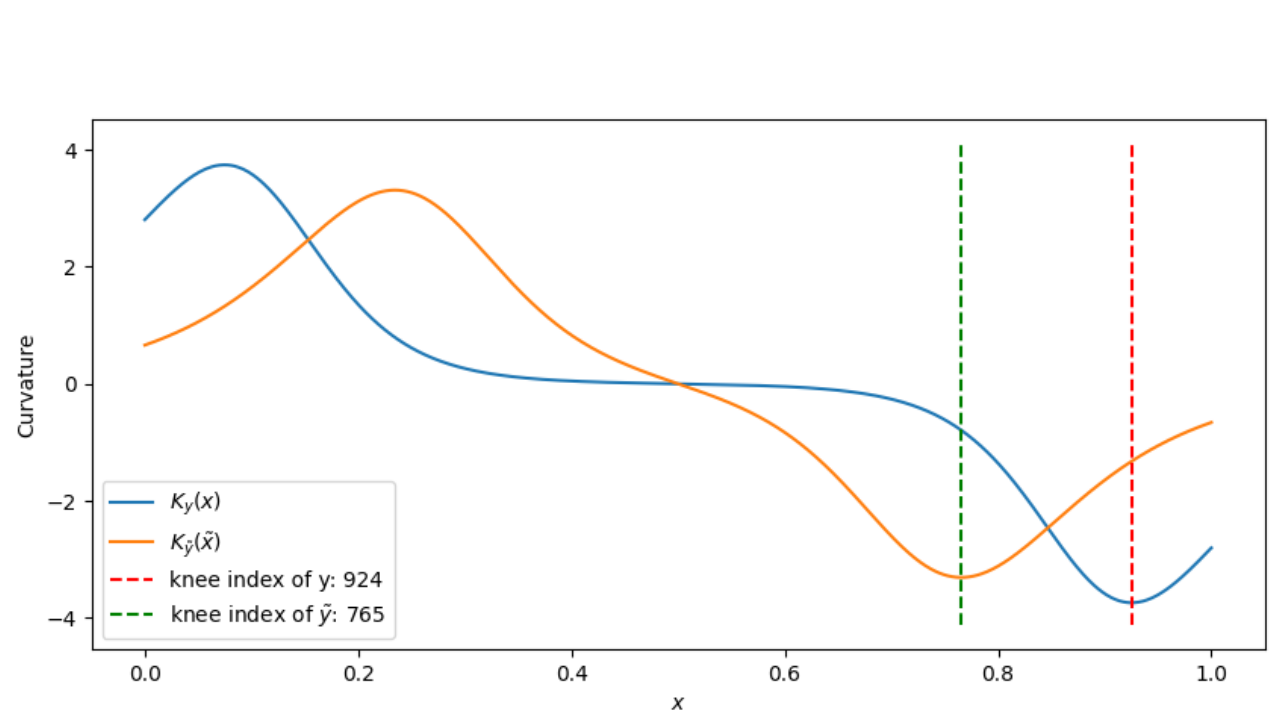}
    \subcaption{}
    \label{fig:sig_norm}
  \end{subfigure}
  \caption{An example showing data normalization changes the curvature shape and knee position. (a) The curve of $y =  5 \times \frac{1}{ 1 + e^{-10x + 5}} $ generated by 1000 evenly-spaced $x$ values in $[0, 1]$. The normalized values are plotted as $\tilde{y}$ in the figure; (b) Curvatures and the corresponding knee point indices of the curves. The normalization operation applies a squeezing effect to the curve of ${y}$, resulting in a smaller rate of change as observed in $\tilde{y}$. This reduces the range of values of $K_{\tilde{y}}(\tilde{x})$ and causes a shift in the position of the knee point.}
  \label{fig:sig_org_and_norm}
  \hspace*{\fill}%
\end{figure*}


In the most common practice, researchers typically use a rule-of-thumbs approach. This intuitive and heuristic method involves plotting the graph and identifying the knee point(s) by visual inspection. An example of this practice can be seen in \cite{PVAD}, where a Partial-Value Association Discovery Algorithm (PVAD) was developed to discover relations in mixed-type real-world data. The first step of the PVAD algorithm requires converting the numeric values of each continuous attribute into categorical values. The technique used in the paper is a heuristic process that involves pinpointing the most substantial jumps in value differences between consecutive data points (known as elbow points) to form data clusters (intervals). However, this ad hoc approach has two main drawbacks: it is highly subjective and the determination of knee points is non-repeatable. Another commonly used method is to define a metric based on system-specific or operational characteristics, which requires prior knowledge. It must be pointed out that system-specific approaches are not practical in the scenario that the dataset being analyzed contains attributes from various domains. 

Our hypothesis regarding this knee point detection problem aligns with the authors in \cite{kneedle}. The aurthors state that a knee point estimation method should be: “Not require tuning for a specific system or operational characteristics is applicable in a wide range of settings”. It is worth noting that the first formal definition of a knee point was documented in that same paper. On top of that, we would like to make an additional assumption that the identification of a knee point should be independent of the data unit, and thus we propose a new definition of knee point in this paper. For the above reasons, we are interested in developing a concise and reproducible knee point determination process. Our objective is to develop a technique that requires minimal human intervention, which can help to identify the data point(s) for grouping data values into intervals that also capture the distribution of data values.

Our main contributions are: i) we provide a novel mathematical definition of knee/elbow, ii) develop a benchmark dataset that includes ground truth of knee point and synthetic samples, iii) propose a new deep learning approach that supports multiple knee point detection, and iv) compare our method with existing methods.

The rest of the paper is structured in the following way. We begin by reviewing related work in Section \ref{sec:related_work}. We then provide a formal definition of knee point in discrete data in Section \ref{sec:knee_point_definition}. Section \ref{sec:proposed_approach} presents our proposed method and  the architecture of our network. The details of the experiment implementation and results are described in Section \ref{sec:experiments}.

\begin{figure*}[h]
  \includegraphics[width=\textwidth,height=6cm]{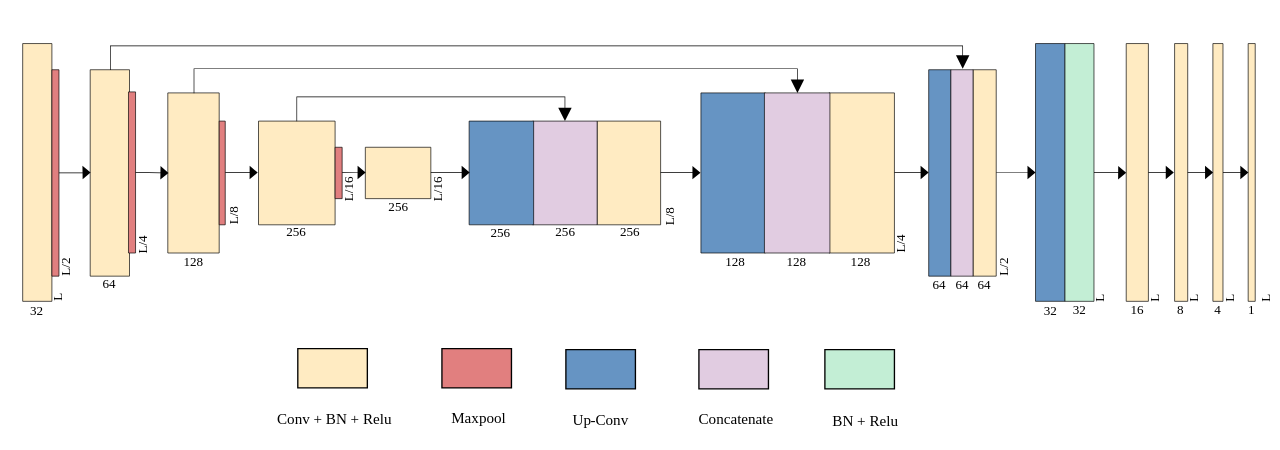}
  \caption{An illustration of the architecture of our proposed method, \textit{UNetConv}. The model is comprised of two main components: a U-Net model and a sequence of convolutional layers. The U-Net model component part passes the input through the encoding path, followed by a bottleneck layer and then to the decoding path. Both the encoding path and decoding path contain four levels of blocks. The numbers beneath and in the bottom right corner of each block respectively indicate the number of channels and size of the resulting feature map passed through that specific layer.}
   \label{fig:model_arch}
\end{figure*}

\section{Related Work}
\label{sec:related_work}

Various approaches \cite{DFDT, AL_and_S_method, Kneedle-8, Kneedle-13, Kneedle-14} have been proposed to identify knees/elbows in discrete data. In this section, we present the most commonly used approaches and compare them in Section \ref{sec:experiments}.

\subsection{\textit{L}-Method}

For every point on the curve except the endpoints, the \textit{L}-Method \cite{Kneedle-8} selects a candidate point and fits a line from the first data point to a candidate point, and fits another line from the candidate point to the endpoint. Root Mean Squared Error (RMSE) is then computed for measuring how close the fitted lines are to data points. The candidate point with the lowest score is selected as an elbow point. However, this method performs best when the size of data points on each side of the elbow is reasonably balanced. It has the tendency to predict a larger elbow index for curves with long tails (more data points on the right side). To overcome this issue, the authors also proposed an iterative refinement method to cut the curve tail and reduce the focus region in each iteration. In each iteration, one candidate elbow point is selected each time. This process stops until the selected elbow value converges. Since this method is designed primarily for determining the ideal number of clusters in cluster analysis, it is only effective when dealing with simple concave curves or curves with limited data size nonetheless. One needs to check the curve shape beforehand and decide the method to be deployed.

\subsection{Dynamic First Derivative Threshold}

The Dynamic First Derivative Threshold (\textit{DFDT}) method \cite{DFDT} is designed to determine the ideal number of clusters in an evaluation criteria plot. It first approximates the first derivative of a curve, followed by using a threshold algorithm (\textit{IsoData}) \cite{DFDT-20} that computes the threshold value for separating the first derivative approximation values into a higher-value and a lower-value group. The elbow point is selected as the data point whose derivative value is closest to the threshold value. One major drawback of \textit{DFDT} is that it is prone to curve with a nearly vertically straight line at the beginning (long curve head), causing the threshold algorithm always returns a larger threshold value. This in turn misleads the method into predicting the elbow point close to the curve head. As such, the authors have incorporated an iterative refinement to the method, similar to the one in the conventional \textit{L}-Method. Instead of removing the curve tail, \textit{DFDT} removes a small segment from the head of the curve in each iteration, specifically the portion from the origin to half of the distance from the previously selected elbow. This process is repeated until the selected elbows converge at the same point. 


\subsection{\textit{AL}-Method}

The \textit{AL}-Method \cite{AL_and_S_method} is an extension of the traditional \textit{L}-method.  The method attempts to determine the point with a sharper angle as an elbow point, so it considers an additional angle score while selecting an elbow point. The angle score is computed as the square of the deviation of the angle between the fitted lines ($\theta_{i}$) from 90 degrees: $|90 - \theta_{i}|^{2}$. Same as the \textit{L}-Method, it requires using linear regression to fit two straight lines for every point on the curve, except the endpoints. The RMSEs and the angle scores are then respectively rescaled to a range of $[0, 1]$,  and combined to calculate the overall score. The point with the lowest score is selected as an elbow point. The authors also deployed the same tail-cutting iterative refinement method to address curves with long heads or long tails, as in \textit{DFDT}. 


\subsection{\textit{S}-Method}

The authors of the \textit{AL}-Method also proposed the \textit{S}-Method \cite{AL_and_S_method} as a further development in the same paper. This method fits three straight lines to a curve to handle curves with long heads or tails. The first and third lines fit for the curve head and tail respectively, while the middle fitted line captures the nature of the curve shape and thus is able to detect the elbow point. The criterion for selecting an elbow point is the weighted RMSE score, which is weighted by the number of data points in the corresponding line segment used in curve fitting. The authors found that using linear regression to fit the points in the selected range introduces bias to the slope of the fitted line. They suggested fitting the lines by using the first and last point in the range but left this as an area for future exploration. This observation is applicable to the \textit{AL}-Method as well.


\hfill

The above-mentioned methods are primarily designed to meet the needs of detecting elbows in clustering applications. The common drawback of these methods is that they are only effective for a narrow range of the $x$-interval (expected number of clusters). Experiments show that these methods have low accuracy when the number of expected clusters is large. Another limitation is that these methods have only been tested on curves with a single elbow point.  If the data points are divided into smaller regions and these methods are called recursively and applied to those regions, non-elbow points may also be incorrectly identified as elbow points. Consequently, these methods do not perform well when used recursively. While \textit{AL} and \textit{S}-Methods have demonstrated excellent performance on error curves from specific clustering algorithms, this indicates that the accuracy of the models can be affected by the underlying clustering algorithm. Therefore, we believe it is crucial to develop a method that is independent of any underlying algorithms and conduct experiments that can test the method's true ability to detect knee points. Nonetheless, both \textit{AL} and \textit{S} methods have a high computational cost, which can be prohibitively expensive for curves with a large number of data points.


\subsection{\textit{Kneedle}}
\textit{Kneedle} \cite{kneedle} is the only algorithm that is capable of detecting multiple knees without the need for recursive calls. The algorithm works by first fitting the data to a smoothing pine, which reduces noise and in an attempt to preserve the original curve shape. The $(x, y)$ values are then normalized into a unit square. After projecting the smoothed points to $y=x$, the method defines a unique threshold value for each local maximum point and determines those local maxima meeting certain conditions as knee points. The rationale behind this is that knees are points further from a straight line. The threshold value is based on the distance between consecutive $x$-values and a user-specified sensitivity parameter $\zeta$. A lower $\zeta$ value tends to declare a knee point more aggressively, which can increase the risk of false positives.  The major weakness of the method is that the fitted smoothing spline may return data points that fall outside the original data range and return irrelevant results.

\subsection{U-Net}

U-Net \cite{unet} is a popular convolutional neural network for biomedical image segmentation. Its architecture consists of successive downsampling and upsampling layers that enable it to learn global features. The network also includes skip connections that can pass local features learned in the same level of the downsampling layer to the upsampling layer at the same level. These local features are then combined with spatial information learned through a sequence of upsampling layers to yield more precise segmentation. Because of its impressive performance in capturing both local and global contextual information of the input image, U-Net has been modified and successfully applied to other visual computing domains such as medical image reconstruction \cite{u_net_app_reconstruct, u_net_app_reconstruct2, u_net_app_reconstruct3, u_net_app_reconstruct4} and pansharpening \cite{u_net_app_pansharpen1, u_net_app_pansharpen2}.

\section{Knee Point Definition}
\label{sec:knee_point_definition}

As in previous works \cite{kneedle, Kneedle-8, DFDT, AL_and_S_method}, a mathematical definition of curvature for continuous function has been used as a foundation for knee/elbow definition. For a twice-differentiable function $f(x)$, the signed curvature of $f$ at point $(x, f(x))$ is given by:

\begin{equation}
  K_{y}(x) = \frac{y''}{ {\left(1 + {(y')}^{2} \right)} ^{3 / 2}}
  \label{eq:curvature_definition}
\end{equation}

Curvature measures the amount by which the tangent vector of the curve changes as the point moves along the curve. The notion of selecting the point of minimum curvature as the knee point is well-suited to heuristics, as minimum curvature captures the exact point at which the curve reaches a peak and then stabilizes instead of continuing to increase or decrease, and as a result, can be used to identify knees. It is noteworthy to mention that in the case of single knee/elbow detection, the problem of finding knee point or elbow point is interchangeable. If a curve presents positive concavity, it can be inverted to a negative concavity curve by replacing the $\mathbf{x}$ and $\mathbf{y}$ data points with the difference of the corresponding maximum value to the original data values (i.e. replace $x_{i}$ by $x_{\text{max}} - x_{i}$ and $y_{i}$ by $y_{\text{max}} - y_{i}$), where $x_{\text{max}}$ and $y_{\text{max}}$ are the maximum values of $\mathbf{x}$ and $\mathbf{y}$ respectively.


However, the above curvature definition is limited to continuous functions, it is not well-defined for discrete data sets. Fitting a continuous function to a set of noisy data is one possible way to extend the definition of curvature on discrete data. Despite the difficulty of fitting, the point identified in the fitting curve may fall outside the valid data range or shift the true knee point position, leading to irrelevant or inaccurate results.

Nevertheless, our goal is to develop an algorithm that performs effectively for datasets having different ranges of values. This is important because real-world data can have a wide range of possible values, and it is crucial for our algorithm to be reliable irrespective of data magnitude. To achieve this, it is necessary to normalize the data into a unit square beforehand.

Let $ D^{N} = \left\{ (\mathbf{x}^{1}, \mathbf{y}^{1}), \cdots, (\mathbf{x}^{N}, \mathbf{y}^{N}) \right\} $ be a set of $N$ samples,  where the $i$-th sample $(\mathbf{x}^{i}, \mathbf{y}^{i})$ consists of $L$ data points such that $\mathbf{x}^{i} = (x^{i}_{1}, \cdots, x^{i}_{L})$ and $\mathbf{y}^{i} = (y^{i}_{1}, \cdots, y^{i}_{L}) $. The rescaling operation that normalizes $(\mathbf{x}^{i}, \mathbf{y}^{i})$ to $( \mathbf{\tilde{x}}^{i}, \mathbf{\tilde{y}}^{i} )$ is: 

\begin{subequations}
\begin{align}
    & { \tilde{x} }^{\,i}_{j} = \frac{ x^{\,i}_{j} - x^{\,i}_{min} }{ x^{\,i}_{max} - x^{\,i}_{min} } \label{eq:norm_x_to_xhat} \\
    & { \tilde{y} }^{\,i}_{j} = \frac{ y^{\,i}_{j} - y^{\,i}_{min} }{ y^{\,i}_{max} - y^{\,i}_{min} } \label{eq:norm_y_to_yhat} 
\end{align}
\end{subequations}
for $j=1, \ldots, L$, where $x^{i}_{min} = min( x^{i}_{1}, \cdots, x^{i}_{L} )$ and $y^{i}_{min} = min( y^{i}_{1}, \cdots, y^{i}_{L})$. The values of $\mathbf{\tilde{x}}^{i}$  and $ \mathbf{\tilde{y}}^{i}$  both fall in the range of $[0, 1]$. If we re-arrange Eq. \ref{eq:norm_x_to_xhat} and \ref{eq:norm_y_to_yhat}, then we have

\begin{subequations}
\begin{align}
    x^{\,i}_{j} & = { \tilde{x} }^{\,i}_{j} ( x^{\,i}_{max} - x^{\,i}_{min} ) + x^{\,i}_{min} \label{eq:norm_xhat_to_x} \nonumber \\
    & : = a^{\,i}_{x} \, x^{\,i}_{j} + b^{\,i}_{x}  \\
    y^{\,i}_{j} & = { \tilde{y} }^{\,i}_{j} ( y^{\,i}_{max} - y^{\,i}_{min} ) +  y^{\,i}_{min} \label{eq:norm_yhat_to_y} \nonumber \\
    & : = a^{\,i}_{y} \, y^{\,i}_{j} + b^{\,i}_{y} 
\end{align}
\end{subequations}

Applying the above results to Eq. \ref{eq:curvature_definition}, the resulting curvature equation of normalized data becomes: 

\begin{equation}
    K_{ \tilde{y} } ( \tilde{x} )  = 
        \frac{ 
            \frac{ a_{x} ^ {2} }{ a_{y} } f''( a_{x} \, \tilde{x} + b_{x} ) 
        }{
            { \left[ 1 + { \left( \frac{a_x}{a_y}  f'(a_{x} \, \tilde{x} + b_{x})  \right) } ^ {2} \right] } ^ { 3/2 }         
        } \label{eq:norm_curv}  
\end{equation}

It is important to note that normalizing data does change the curve shape and thus alter the knee/elbow point position. Fig. \ref{fig:sig_org} and \ref{fig:sig_norm} demonstrate how normalizing data changes the curvature shape and knee position. Fig.\ref{fig:sig_org} shows the curve of $y = 5 \times \frac{1}{ 1 + e^{-10x + 5}} =  5 \times \tilde{f} (x)$ generated by $L=1000$ data points, where $\mathbf{x} = (x_{1}, \cdots, x_{1000})$ are values evenly-spaced in the interval $[0, 1]$ and each corresponding $y_{j}$ is calculated by $y_{j} =  5 \times \tilde{f} (x_{j})$. The curvature of the function is $K_y(x) = 
\frac{
    5 \cdot 10 \cdot \tilde{f} (1 -\tilde{f}) (1 - 5\tilde{f})
}{
    \left[ 1 + 
        \left( 5 \cdot 10 \cdot \tilde{f} (1 -\tilde{f}) \right)^{2}  
    \right] ^ {3/2}
} $. By inputting the corresponding values of  $x_{min}, x_{max},  y_{min}, y_{max}$ which are 0, 1, 0.033, and 4.967 into Eq. \ref{eq:norm_curv} and making the necessary substitutions, we have $
    K_{ \tilde{y} } ( \tilde{x} )  = 
        \frac{ 
            \frac{ 1}{ 4.934 } f''( \tilde{x}) 
        }{
            { \left[ 1 + { \left( \frac{1}{4.934}  f'( \tilde{x})  \right) } ^ {2} \right] } ^ { 3/2 }         
        }
$. In Fig. \ref{fig:sig_norm}, one can observe that the curvature of $\mathbf{y}$ ranges from -3.740 and 3.740, while that for $\mathbf{\tilde{y}}$ is -3.308 to 3.308. The change in curvature value is a result of the rescaling operation applied to the $x$ and $y$ values. This operation compressed the 1000 $x$ and $y$ values into shorter intervals of $[0, 1]$ respectively. As a result, the curve of $\mathbf{\tilde{y}}$ is flatter than $\mathbf{y}$, resulting in a slower deviation of the tangent to the curve of $\mathbf{\tilde{y}}$ in the interval. This in turn leads to a decrease in curvature values and a shift of the knee point to a forwarder (leftward) position.

\begin{table*}[t]
  \centering
   \begin{tblr}{|c|c|l|c|} 
    \hline 
    Code & Function & Description & Flipped?  \\
    \hline 
    
    \textit{FT1} & $\ln(x)$
        &  Logarithm   
        & Y \\    
   \hline
    \textit{FT2} & $(-1)^{m + 1} x^{m}$, for $m = 3, 5, 9$ or $11$ 
        & Polynomial 
        & Y \\
   \hline
    \textit{FT3} & $x^{ \frac{1}{m} }$, for $m = 3, 5, 9, \ldots, 17$ 
        & Rational
        & Y \\
    \hline
    \textit{FT4} & $\frac{1}{ 1 + e^{-x}}$ 
        & Logistic
        & Y \\      
    \hline
    \textit{FT5} & $-\ln{(1 + e^{-x})}$ 
        & Translated Softplus
        & Y \\
    \hline
    \textit{FT6} & $ 1 - e^{- x}$ 
        & Translated SELU
        & Y \\
    \hline
    \SetCell[r=2]{c} \textit{FT7} & 
    \SetCell[r=2]{c} $(\frac{mx}{s})^{p} - (\frac{mx}{s})^{q} e^{-(\frac{x}{s})^{r}} $ &
    Product of exponential and  & 
    \SetCell[r=2]{c} N \\
    
     & & rational function & \\
    
    \hline
    \textit{FT8} 
    & $y(x) 
        =\begin{cases}
              m_{1} x  &, \text{if} \ x \in [0, \ x[\text{knee\_index} - 1] ] \\
              m_{2} x + c_{2}  &, \text{if} \  x \in [x[\text{knee\_index}], \ 1] 
        \end{cases} $
    & Pieceswise Linear
    &  N \\
    \hline
    \textit{FT9}
    & $ \frac{1}{\sigma \sqrt{2 \pi}} \int^{x}_{-\infty} e^{- \frac{(t - \mu)^{2}}{ 2 \sigma^{2} }} \quad , \text{where} \ \mu = 13, \sigma=5$
        & Normal Distribution CDF
        & Y \\ 
    \hline
    \textit{FT10} & $\sum_{i=1}^{ K } \frac{c_{1, i}}{ 1+ e^{-c_{2, i} (x - c_{3, i})}}$ 
        & Sum of $K$ Logistic functions
        & N   \\  
    \hline
    \rule{0pt}{10pt}
    \textit{FT11} & 
    $ \frac{1}{m}
        \sum_{i=1}^{K} 
        \frac{ \binom{2K}{K - i} }{ i \cdot 2^{2K - 1} 
        } \sin(ix)
        + (x + t) \cdot q \cdot \ln(x)
    $ 
    & Translated tilted sine
    & N 
    \\ 
    \hline

    \textit{FT12} & 
    Sum of \textit{FT1-FT8}
    & Sum of single-knee functions
    & N 
    \\
    \hline
  \end{tblr}
  \caption{Selected functions to generate samples and create datasets. If an inverse function exists, the normalized samples generated by that function will be randomly flipped along $y=1-x$. For \textit{FT7}, the values of $p, q, r$ are determined by a brute force search in the set of values $[1, 2, 3, 4, 5]$, the possible values of $s$ are $\{10, 20\}$ and that for $m$ is $\{0.1, 0.2, \ldots, 5.0\}$. For \textit{FT10} and \textit{FT11}, $K$ is the number of knees in a sample.}
  \label{tab:data_true_distribution}
\end{table*}

\section{Proposed Approach}
\label{sec:proposed_approach}

This section describes the architecture of our proposed convolutional neural network  (\textit{UNetConv}) and introduces the loss function and inference method used for knee point detection.

\subsection{Model Architecture}

The proposed network architecture is displayed in Fig. \ref{fig:model_arch}. It should be pointed out that the height and width of each layer output are not entirely drawn to scale. The model is comprised of two main components: a U-Net model and a sequence of convolutional layers. The first part receives the input and processes it through the encoding path, then through a bottleneck, and finally through the decoding path. Both paths consist of four levels of blocks. In the encoding path, each block has a convolutional layer with $11 \times 11$ kernel with same padding, followed by a batch normalization (BN) layer and a ReLU activation function. The last layer of each block is a $2 \times 2$ max pooling layer with a stride of 2, which reduces the feature map width by half for the purpose of downsampling. The bottleneck layer consists of a single convolutional layer, which also has an $11 \times 11$ kernel, same padding, and 256 channels.

In the decoding path, additional layers are added to every level. First, an up-convolutional layer with $2 \times 2$ kernel is applied to upsample a feature map, followed by a BN layer and a ReLU activation function. A skip connection then takes place, which concatenates the feature map with the output from the encoding part at the same level. Lastly, the feature map is fed to a convolutional layer of $11 \times 11$ kernel with same padding. In both encoding and decoding paths, the number of channels in each block is 32, 64, 128, and 256. 

The second part of the model is a sequence of convolutional layers. Each layer has is a $2 \times 2$ convolutions and same padding. The number of channels in the layers is 16, 8, 4, and 1 respectively. The final step is normalization,  which maps the output to a probability, indicating the likelihood of the current data point being a knee point. The network contains about 3.3M parameters in total.

\subsection{Soft $F_{1}$ score}
\label{subsec: soft_f1}

The $F_{1}$ score is a statistical measure  to evaluate the accuracy of a classification model. It is particularly useful when the classes/labels in the dataset are imbalanced, which is the case of our scenario - there are at most 5 knee points in each sample. The traditional $F_{1}$ score is a harmonic mean of two other metrics - precision and accuracy. Details are discussed in Section \ref{sec:metric}.

The issue with the traditional $F_{1}$ score is that it is not differentiable. It accepts binary (0 or 1) inputs of prediction and ground truth,  then it computes integer values of True Positive, False Positives, and False Negatives. Thus, it can not be used as a loss function to compute the gradient and update the model's weights during the training phase. This limitation can be overcome by modifying the $F_{1}$ metric to accept probabilities as inputs and calculate the required counting numbers as a continuous sum of likelihood. The equation is given in the following.

Let $(\hat{p}^{i}_{1}, \cdots, \hat{p}^{i}_{n} )$ be a set of predicted probabilities by the model and $(p^{i}_{1}, \cdots, p^{i}_{n}$) be the binary ground truth of sample $i$ respectively, where $p^{i}_{j} = 1$ if a knee is attained at index $j$, 0 otherwise. The soft F1-score $\tilde{F}_{1}$ is :

\begin{equation}
    \tilde{F}_{1}(  \mathbf{\widehat{p} }^{i}, \mathbf{p}^{i} ) = 
    \frac{
        \sum_{j=1}^{L} \hat{p}^{i}_{j} \ p^{i}_{j}
    }{
        \sum_{j=1}^{L}\hat{p}^{i}_{j} + \sum_{j=1}^{L} {p}^{i}_{j}
    }
    \label{eq:soft_f_score}  
\end{equation}

$\tilde{F}_{1}$ naturally approximates the traditional $ F_{1}$ classification metric and shares the same property of indicating better precision and accuracy with a higher score (with 1 being the best value).

\subsection{Non-maximal Suppression}

Non-maximal suppression (\textit{NMS}) \cite{non-max-suppression} is a common technique in computer vision to eliminate multiple detections of the same object. With a pre-specified threshold value $\delta$ and the area of interest (suppression area), the process of \textit{NMS} involves dropping all detections whose prediction values are below $\delta$. The algorithm then selects the highest-scoring candidate and suppresses all other overlapping candidates within the area of interest. This process repeats until no candidates remain.

In this research, the \textit{UNetConv} model is designed to predict the probability of a data point being a knee point. Often it is clear which data point is a knee and the probability curve predicted by the model presents a tall narrow spike shape. However, there are cases where multiple spikes may occur near a knee point, which could be due to noise or the model naturally predicting higher probability near a knee point, making it difficult to determine. For this reason, we implement \textit{NMS} to fix this issue.

\section{Experiments}
\label{sec:experiments}

This section explains how we create noisy data for the training and test sets. We then provide details on how we implement the experiment. Finally, we evaluate the proposed network and compare it to other existing methods.

\subsection{Synthetic Data}
\label{subsec:data_collection}

 To evaluate the model performance, we select twelve functions to generate samples and create datasets. These functions (\textit{FT1-FT12}) are listed in Tab. \ref{tab:data_true_distribution}. One important note here is that we make certain assumptions about the curve being used. Specifically, we assumed that the curve has two main characteristics: (1) monotone
increasing, and (2) having at least one knee point in the interval. \textit{FT1-FT9} are functions that have only one knee point in the interval. \textit{FT6} is the translated Scaled Exponential Linear Unit (SELU) function and \textit{FT9} is the cumulative distribution function (CDF) of normal distribution. On the other hand, the number of knees $K(\geq 2)$ for \textit{FT10-FT12} samples can be specified, which allows for multiple knee points in these functions. In these multi-knee functions, \textit{FT10} is the sum of multiple logistic functions and its curve is a smooth step function. \textit{FT11} is a combination of sine functions, the resulting function can be described as a translated tilted sine function. \textit{FT12} (see Fig. \ref{fig:FT12sample}) is a synthetic function formed by the summation of functions from the single-knee family. The number of functions chosen to generate a sample of \textit{FT12} corresponds to the number of knees in the sample. Each time, a function from the single-knee family is randomly selected to concatenate with the currently connected curve. There is one restriction on the slope when joining the curves: the slope formed by the last two points in the existing curve should not be greater than the slope of the first two points of the next curve being connected. This prevents the creation of additional knee points upon connection. 

To introduce some noise to the generated sample, while maintaining the function data range, we consider each function in Tab. \ref{tab:data_true_distribution} as a cumulative distribution function (CDF). We then generate noisy data points $\mathbf{\hat{y}}$ by making use of the empirical distribution function and Inverse Transform Sampling method. Mathematically, given a CDF $f_{\widetilde{X}}$ and a uniform variable $U \sim Uniform[0, 1]$, the random variable $R = f_{\widetilde{X}}^{-1}(U)$ can be described by $f_{\widetilde{X}}$. Therefore, the empirical distribution of $R$ can be written as the following: 

 \begin{align} 
 \widehat{f}_{R}(r) 
  & = \frac{
    \sum_{j=1}^{L'} \mathbf{1}_{R_{j} \, \leq \, r}
    }{L'} 
    & \nonumber  \\
  & = \frac{
    \sum_{j=1}^{L'} \mathbf{1}_{f^{-1}(U_{j}) \,  \leq \,  r}
    }{L'}  
    & \left( \because R_{j} = f_{\widetilde{X}}^{-1}(U_{j}) \right) \nonumber  \\
  & = \frac{
    \sum_{j=1}^{L'} \mathbf{1}_{U_{j} \, \leq \,  f(r)}
    }{L'}  
    & 
 \label{eq:empirical_cdf}
  \end{align}  
 , for $r \in \mathbf{\widetilde{X}}$. Based on the above results, we can obtain noisy data points $\mathbf{\hat{y}}^{i}$ by using the cumulative count of randomly generated points that follow a standard uniform distribution with a value less than $\tilde{y}_{j}$. The complete procedure to generate a sample is summarized in the following:

\begin{figure}
\includegraphics[width=0.85\textwidth]{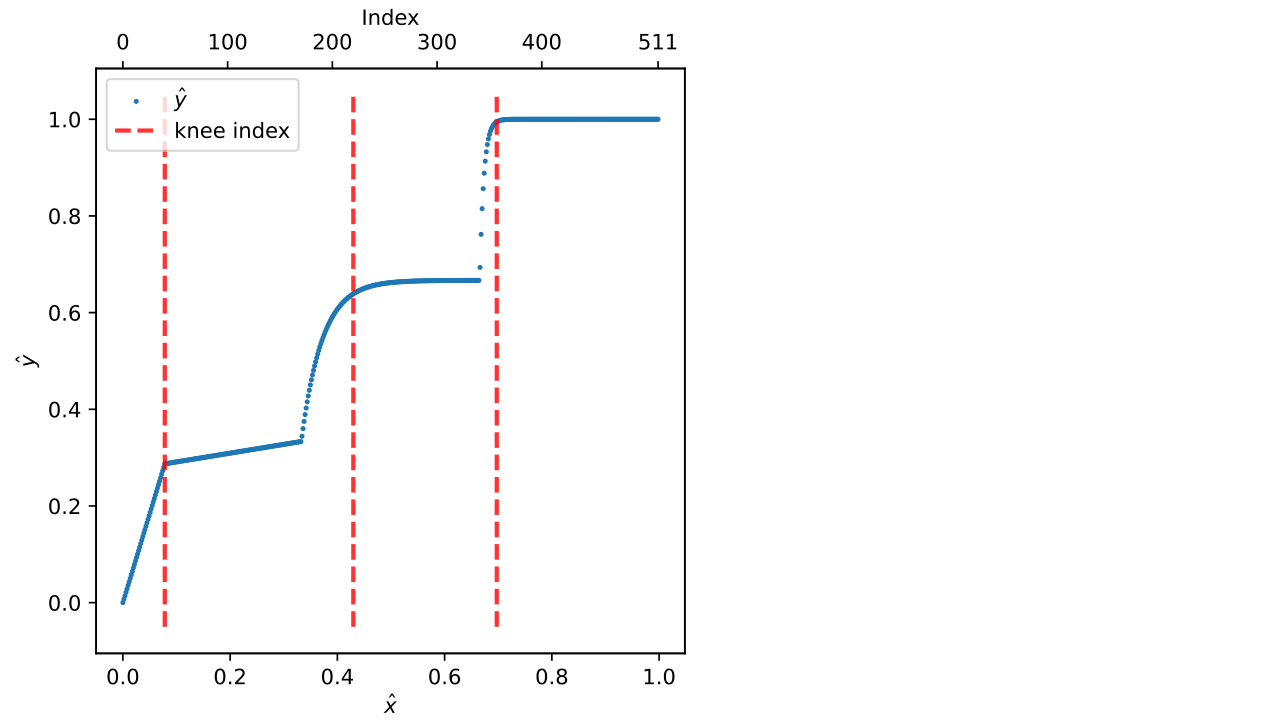}
\caption{
    A graphical representation illustrates a \textit{FT12} multi-knee sample, created by summing the graphs of individual single-knee function. This composite sample is created in the sequence of \textit{FT8}, \textit{FT1}, and \textit{FT6}.
    }
\label{fig:FT12sample}
\end{figure}

\begin{enumerate}[label=\roman*.] \label{list:step_to_gen_noisy_data}
    \item Generate $L$ pairs of noise-free data points $(x^{\,i}_{j}, y^{\,i}_{j}) $ where 
    \begin{align*}
        x^{\,i}_{j} &= x^{\,i}_{lb} + \frac{j}{L-1} (x^{\,i}_{ub} - x^{\,i}_{lb})\\
        y^{\,i}_{j} &= f(x^{\,i}_{j})
    \end{align*}
    for $j = 0, 1, \cdots, L-1$ 
    \item Compute the normalized values $( \mathbf{\widetilde{x}}^{i}, \mathbf{\widetilde{y}}^{i} )$ of sample $i$ by inputting the results from the previous step into Eq. \ref{eq:norm_x_to_xhat} and \ref{eq:norm_y_to_yhat}
    \item Generate $\{ u_{m} \}_{m=1}^{L'}$  from the standard uniform distribution $Uniform[0, 1]$, where $L'$ is not necessarily same as $L$

    \item Compute the noisy data points $(\hat{x}^{\,i}_{j}, \hat{y}^{\,i}_{j})$ for sample $i$, where
    \begin{subequations}
        \begin{align}
            \hat{x}^{\,i}_{j} & = \tilde{x}^{\,i}_{j} & \\
            \hat{y}^{\,i}_{j} & = \frac{
                \sum_{m=1}^{L'} \mathbf{1}_{u_{m} \, \leq \,  \tilde{y}^{\,i}_{j}}
                }{L'}   \, 
        \end{align}
    \end{subequations}
\end{enumerate}

\begin{figure*}
  \centering
  \begin{subfigure}{0.48\linewidth}
    \includegraphics[width=500pt]{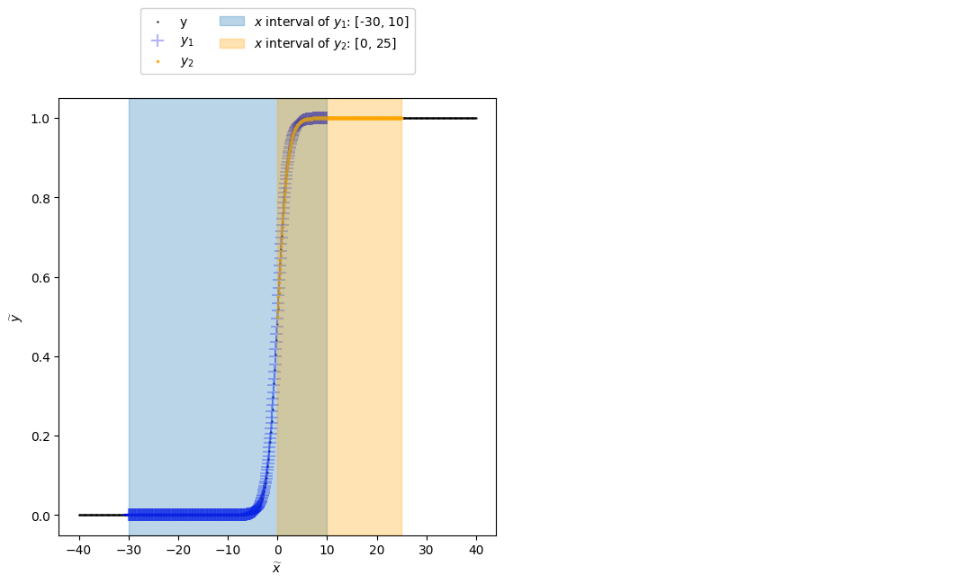}
    \subcaption{}
    \label{fig:sig_in_diff_x_intervals}
  \end{subfigure}
  \hfill
  \begin{subfigure}{0.48\linewidth}
    \includegraphics[width=500pt]{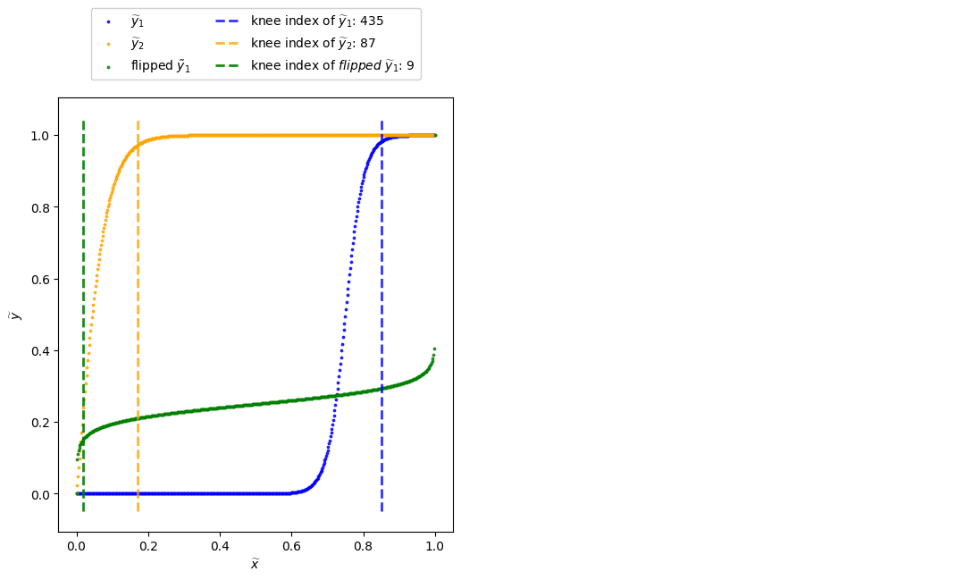}
    \subcaption{}
    \label{fig:sig_norm_in_diff_x_intervals}
  \end{subfigure}
  \caption{An example showing varying the $x$ interval can generate samples with a variety of curve shapes, different ranges of curvature values, and thus different positions of knee point(s). (a) A graph showing the logistic function, $y = \frac{1}{1 + e^{-x}}$, for $x \in [-40, 40]$. (b) The curve shape of $\tilde{y}_{1}$ is noticeably different from $\tilde{y}_{2}$, even though both are produced by the same function. The figure also shows the flipped curve of $y_{1}$. Unlike the logit function, the knee point of \textit{flipped} $\tilde{y}_{1}$ occurs at the very beginning of the curve.}
  \label{fig:sig_diff_shape}
\end{figure*}

It is noteworthy to mention that the $x$ interval varies between samples, even if they are generated from the same function. Generally speaking, a wider $x$ interval leads to a curve with sharper curvature after normalization. Varying $x$ interval allows us to create samples with different ranges of curvature values. Another benefit is that the curve may have different shapes for different intervals of $x$. Taking \textit{FT4} as an example, the curve exhibits an elbow point at approximately $x=-1.36$ (see curve $y_{1}$ in Fig. \ref{fig:sig_in_diff_x_intervals}).

To further improve the diversity of data, we also randomly flip a normalized sample along the $y=1-x$ axis if there exists an analytical expression for the inverse of the chosen function. Again taking \textit{FT4} as an example, its inverse function is the logit function $y = ln (x/(1-x))$ which has a significantly different curve shape and different knee point position, even though the points are generated by the same interval [-30, 10] (Fig. \ref{fig:sig_norm_in_diff_x_intervals}). Fig.  \ref{fig:sig_norm_in_diff_x_intervals} shows the flipped data of $\tilde{y}_{1}$. The knee point of \textit{flipped} $\tilde{y}_{1}$ occurs at the very beginning of the curve.

\subsection{Implementation Details}

\begin{description} [leftmargin=0pt]

\item[Synthetic Training Set and Test Sets]
We create three distinct datasets using the functions outlined in Tab. \ref{tab:data_true_distribution} in order to test the model performance. The network is trained on 7000 samples from \textit{FT1-FT8, FT10-FT12}, reserving \textit{FT9} for building a separate test set since we want to investigate the model's performance on samples from unseen functions. In the 7000-sample training set, there is an equal portion of single-knee and multi-knee samples, with 3500 of each category. Each function is proportionally represented within its respective portion. Specifically, functions \textit{FT10-FT12} each account for 33.33\% in all multiple knee samples, while the remaining distributions each account for 12.15\% of all the single knee samples. The model performance is tested by identifying knee point index/indices in three test sets containing $N=300, 800$, and 100 samples, respectively. The 300-sample test set (denoted as $mknee$) includes 100 samples from each multi-knee distribution (\textit{FT10-FT12}), and the 800 test set (denoted as $sknee$) also comprises 100 samples from each single-knee distribution (\textit{FT1-FT8}). The last test set contains 100 samples from \textit{FT9} distribution (denoted as $ng$). Since the model has not been trained with samples from this distribution, its outcome will demonstrate the model's capability in capturing knee point properties. Every sample in the data sets has $ L= 512$ pairs of $(x, y)$ data points. Upon analyzing the data, the range of knee curvature values per dataset is as follows: [-337.32, -3.00] for the training set, [-339.39, -3.00] for the 800-sample $sknee$ test set, [-326.32, -5.78] for the 300-sample $mknee$ test set, and [-40.85 to -6.82] for the 100-sample $ng$ test set. To sum up, the range of knee curvature in this experiment is between -339.39 and -3.00. In all of the synthetic data sets, the knees are not located within 10 indices from the boundary.

\begin{table}
  \centering
  \begin{tabular}{r|l}
    \toprule
    Configuration & Values \\
    \midrule
    Center of Mass & 0.2, 0.4, \ldots, 10.0 \\
    Span & 1.2, 1.4, \ldots, 10.0 \\
    Half-life & 0.2, 0.4, \ldots, 10.0 \\
    Alpha & 0.1, 0.3, \ldots, 0.9 \\
    \bottomrule
  \end{tabular}
  \caption{Values of configuration that were attempted when applying the Exponentially Weighted Moving Average (EWM) to smooth data. The configuration value that achieves the lowest MSE between the smoothed data and noise-free data is chosen. The optimal configuration varies for each sample.}
  \label{tab:kneedle_smooth_config}
\end{table}

\item[Data Preprossing for Other Model/Methods]
Since the \textit{Kneedle} method requires curve smoothing in the data preprocessing stage, each sample is first smoothed by Exponentially Weighted Moving Average (EWM) before being projected/rotated. We test various configurations (as shown in Tab. \ref{tab:kneedle_smooth_config}) and select the one that achieves the lowest MSE between the true noise-free $\mathbf{y}$ and the fitted curve for each sample. In addition, for methods primarily designed for detecting elbows in clustering applications such as \textit{DFDT} and \textit{AL}-Method, we translate the data points of each single-knee sample by $(\tilde{x}^{\,i}_{j}, \ 1 - \tilde{y}^{\,i}_{j} )$ such that the translated curve has consistent positive concavity, which is similar to the loss function shape in clustering applications.

\item[Loss Function]
As discussed in Section \ref{subsec: soft_f1}, the traditional $F_{1}$ is an intractable step function for gradient descent. To overcome this limitation, we implement the soft $\widetilde{F}_{1}$ score as a surrogate function of $F_{1}$. Strictly speaking, our loss function is defined as:

\begin{equation}
    \min \frac{\alpha}{\widetilde{F}_{1}} +  1 - \widetilde{F}_{1}
    \label{eq: loss_func}
\end{equation}

, where $\alpha$ is a constant. To determine the value of $\alpha$, we run our model with $\alpha$ set to 0.01, 0.1, and 1, each with a single trial, using the loss function described in Eq. \ref{eq: loss_func}. We select the $\alpha$ value that results in the lowest loss value for the subsequent experiment. The selected $\alpha$ value is 0.1.

\item[Optimization]
We train the network for 200 epochs with batch size = 64. AdaDelta is employed with an initial learning rate = 0.5 and momentum = 0.5. For every 10 epochs, the learning rate is decreased by half. The loss function used is given in Eq. \ref{eq:soft_f_score}. 

\item[Post-Processing on Model Output]
In the testing stage, the model output of the $i$-th sample, $\mathbf{\widehat{p}}^{i}$, undergoes further processing using \textit{NMS}. This method predicts the knee index/indices and returns a binary prediction $\mathbf{\widehat{p}}_{NMS}^{\,i}$ with a value of 1 at index $j$, indicating the detection of a knee at the $j$-th data point. The probability threshold selected for \textit{NMS} is 0.5. The suppression area is set to be $\pm$ 10 indices, dropping any candidates located 10 indices from the left and 10 indices from the right of the selected knee point in each iteration. The resulting binary output $\mathbf{\widehat{p}}_{NMS}^{\,i}$ is compared with the binary ground truth $\mathbf{p}^{i}$ to compute the traditional $F_{1}$ score for model performance evaluation.
\end{description}


\subsection{Metric}
\label{sec:metric}

The $F_{1}$ score is one of the most widely used metrics in classification analysis. It is the harmonic mean of precision and recall, providing a single score that balances these two metrics. This is useful because a model with high recall but low precision may correctly identify a lot of true positives, but may also identify many false positives. 

Since some algorithms cannot detect at the exact same knee point, we incorporate allowable index error in the calculation of $F_{1}$ score to accommodate for this issue as in \cite{kneedle}. As an illustration, suppose we have a pair of ($\mathbf{x}, \mathbf{y}$) values with $L$ = 7, and the knee occurs at $i = 5$ and $7$. We consider the algorithm to have "correctly" identified the knees if it identifies any points at $i = 3, 4, 5, 6,$ or $7$ as knees, with a margin of error of $2$.

\subsection{Evaluation}

\begin{table*}[t]
    \centering
    \begin{tblr}{r|ccccccccc}
    \hline
    Test Set  & DFDT & DFDT Ref & AL & AL  Ref & S  & S Ref & Kneedle Proj & Kneedle Rot & UNetConv \\ \hline
    sknee   &   0.021 & 0.024    & 0.274           &  $\bowtie$              & 
    0.044          & $\bowtie$               & 0.09        & 0.09       & \textbf{0.740}    \\ \hline
    mknee   &   --   &    --  &   --  & -- & --  & -- & 0.063        & 0.063       & \textbf{0.720}    \\ \hline
    ng & 0.000 & 0.000    & 0.040           & $\bowtie$                 & 0.000          & $\bowtie$              & 0.11        & 0.11       & \textbf{0.810}    \\ \hline
    \end{tblr}
\caption{Quantitative results of \textit{UNetConv} and other methods, with an allowable index error of 2. The symbol $\bowtie$ denotes that the method failed to converge.}

\label{tab:result_allow_error_2}
\end{table*}

\begin{figure*}
  \hspace*{\fill}%
  \begin{subfigure}{0.33\linewidth}
    \includegraphics[width=330pt]{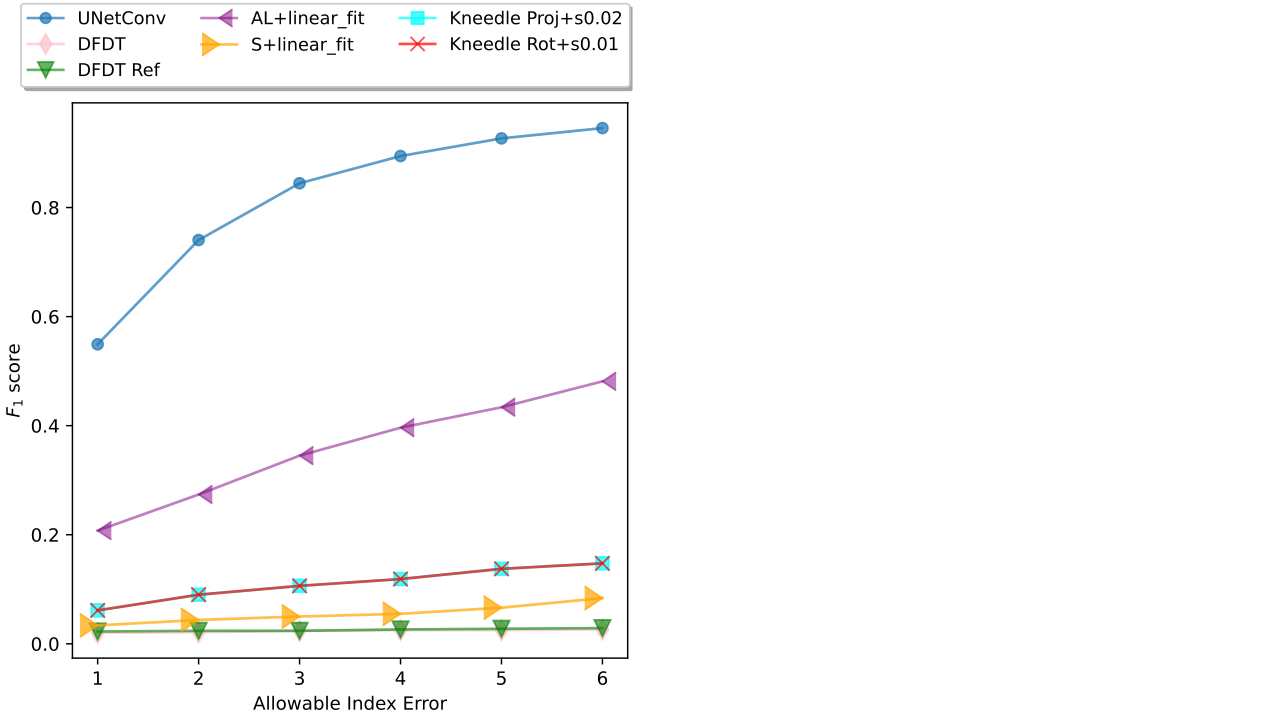} 
    \subcaption{}
    \label{fig:result_sknee}
  \end{subfigure}
  \begin{subfigure}{0.33\linewidth}
    \includegraphics[width=330pt]{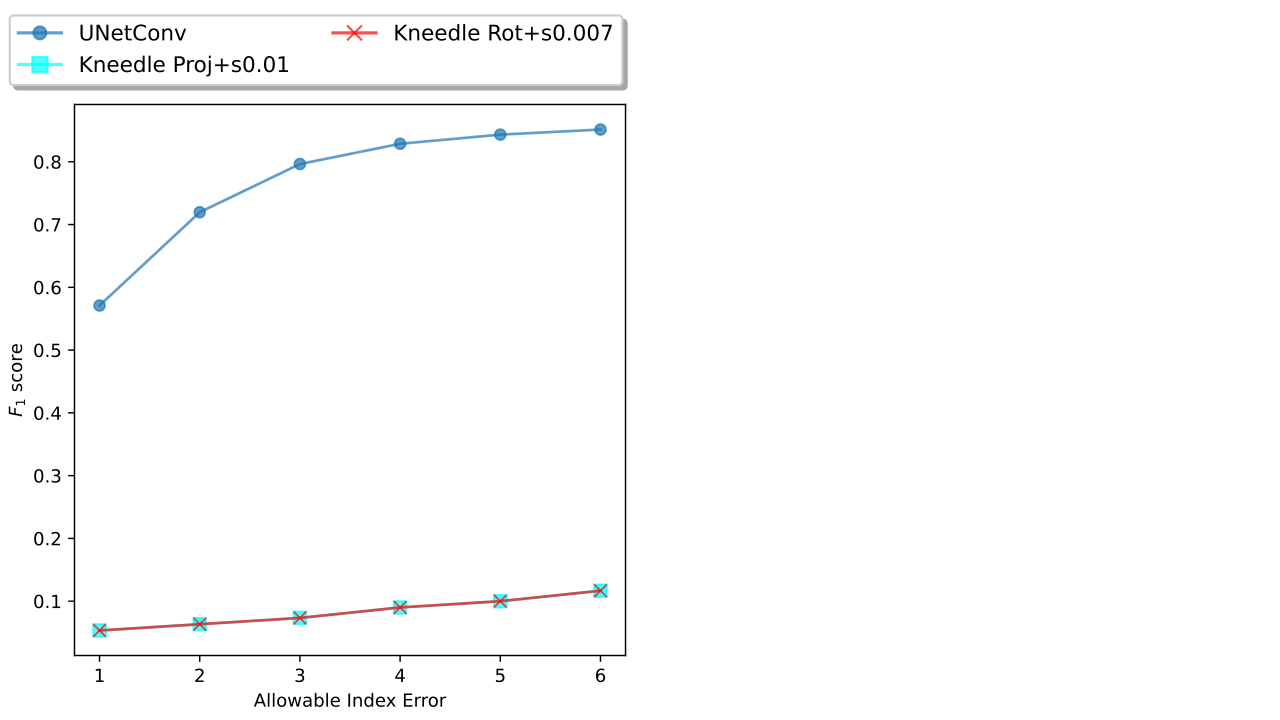} 
    \subcaption{}
    \label{fig:result_mknee}
  \end{subfigure}
  \begin{subfigure}{0.33\linewidth}
    \includegraphics[width=330pt]{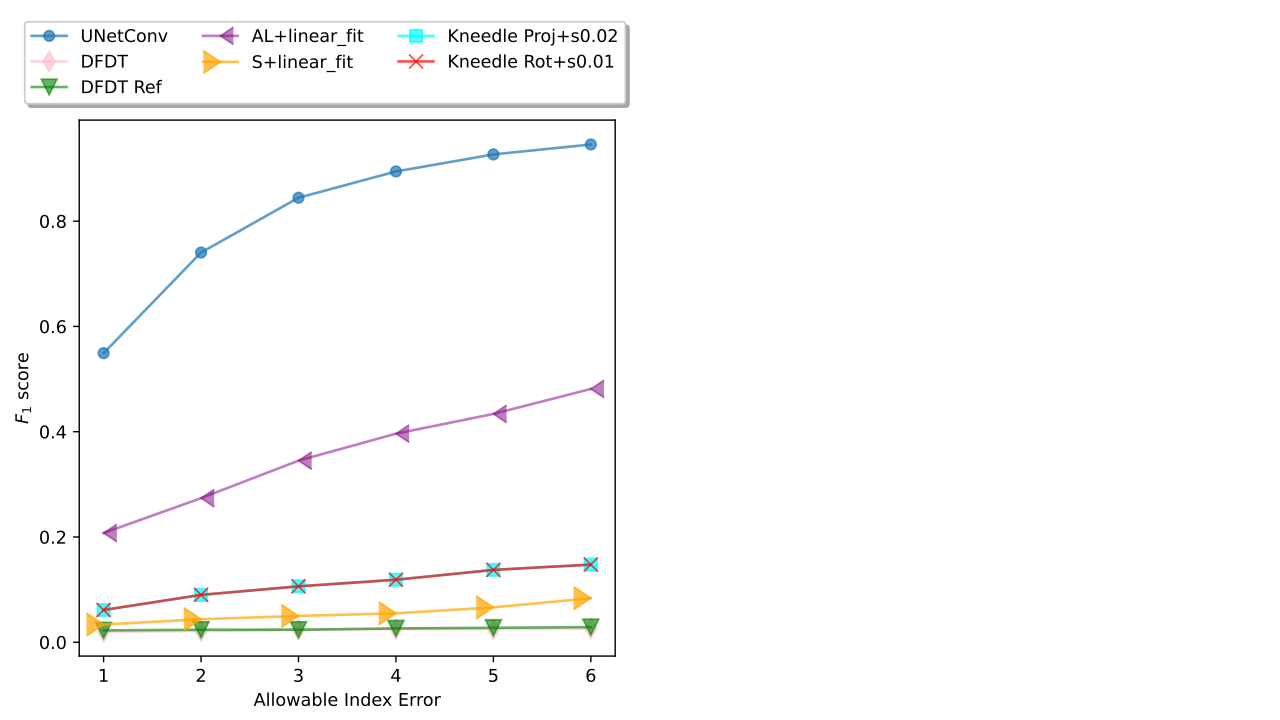}
      \subcaption{}
    \label{fig:result_sig}
  \end{subfigure}
  
  \caption{$F_{1}$ scores of (a) \textit{UNetConv}, \textit{DFDT}, \textit{AL}, \textit{S} and \textit{Kneedle} methods for varying allowable index error on the $sknee$ data set; (b)\textit{UNetConv} and \textit{Kneedle} methods for varying allowable index error on the $mknee$ data set; (c) \textit{UNetConv}, \textit{DFDT}, \textit{AL}, \textit{S} and \textit{Kneedle} methods for varying allowable index error on the $ng$ data set.
  }
  \label{fig:results_testset}
  \hspace*{\fill}%
\end{figure*}

We evaluate our network by running 20 trials on all three synthetic test sets. The average of the $F_{1}$ scores per test set is then compared with other methods mentioned in Section \ref{sec:related_work}. In Fig. \ref{fig:results_testset}, the $F_{1}$ scores for each method per test set are plotted against allowable errors ranging from 1 to 6. Our method is denoted by \textit{UNetConv}. We use two curve-fitting methods for both \textit{AL} and \textit{S} methods: fitting a straight line that best matches all the data within the specified range (\textit{best fit}), and fitting the first and last data point within the specified range (\textit{linear fit}). Since the results using \textit{linear fit} are better than \textit{best fit} irrespective of \textit{AL} or \textit{S} methods in all scenarios, we only list the results using \textit{linear fit} in Fig. \ref{fig:results_testset} and Tab. \ref{tab:result_allow_error_2}.  The iterative refinement methods of the \textit{AL}-Method and \textit{S}-Method are denoted as AL Ref and S Ref, respectively. However, we are unable to obtain results as these methods fail to converge for some samples. In the original work of \textit{Kneedle}, projection is implemented to transform data points instead of rotation. We perform both projection (\textit{Kneedle Proj}) and rotation (\textit{Kneedle Rot}) when comparing the results. For each test set and each data transformation method, we select the $\zeta$ value that achieves the highest average $F_{1}$ score among the allowable index errors and only consider these results when making comparisons. For \textit{Kneedle Rot}, the best $\zeta$ values for $sknee, mknee$ and $ng$ are 0.01, \{0.007, 0.008\} and \{0.006, 0.007, \ldots, 0.02\} respectively. As for \textit{Kneedle Proj}, the corresponding ideal $\zeta$ values for $sknee, mknee$ and $ng$ are 0.02, 0.01, and \{0.008, 0.009, \ldots, 0.03\}. For simplicity, we take the lowest $\zeta$ value since there are multiple $\zeta$ values achieving the same mean $F_{1}$ score. To demonstrate the overall performance of the knee detectors, we plot the knees detected by \textit{UNetConv} and the techniques that achieve the top three $F_{1}$ scores in Fig. \ref{fig:result_sknee}-\ref{fig:result_sig}. Tab. \ref{tab:result_allow_error_2} shows the quantitative results of the proposed model and other methods, with an allowable error of 2. 

\subsection{Discussions on the results}

In all the synthetic test sets and all settings of allowable index error, \textit{UNetConv} outperforms the existing methods. From  Tab. \ref{tab:result_allow_error_2}, our model attains the highest $F_{1}$ score of 0.74 in the $sknee$ test set, with the second best result of 0.274 attained by the \textit{AL}-Method. The results of other methods are all below 0.1. For the multiple-knee $mknee$ test set, \textit{UNetConv} obtains the highest score of 0.72, followed by \textit{Kneedle} (0.063), which results in the same score regardless of using rotation or projection to transform the curve. For the unseen 100-sample noisy gaussian $ng$ test set, \textit{UNetConv} again surpasses other methods. 

The \textit{DFDT} techniques are not capable of locating knee points as most of the samples have both an elbow point and a knee point on the curves. The use of only first derivative values makes it challenging to differentiate between an elbow and a knee. It is thus understandable that the method struggles to have good performance. Though there is an iterative refinement method to help overcome this shortcoming, there is only a minor improvement in the model performance. Based on our analysis of the test sets, we have found that these methods perform better on the \textit{FT8} dataset, which consists of noisy piecewise straight-line functions. This is due to the fact that the first-derivative values of these samples can be distinctly separated into higher and lower value groups, thus making it easier for the method's threshold algorithm \textit{IsoData} to accurately estimate the optimal knee point.

For the \textit{AL}-method, it achieves the second highest $F_{1}$ score on the $sknee$ dataset. This method works best on simple curve shapes, like \textit{FT8} and \textit{FT9}, where each curve segment is close to a straight line and the angle score can accurately capture a knee point. However, when it comes to the curve segment with a quadratic or higher degree of curve shape (as in \textit{FT7}), the fitted straight lines and the angle between no longer contribute positively towards locating a knee. We also observe that this method is not as effective when being used on curves with a gradual rate of change or less sharp curvature.

The \textit{S}-method works poorly on all the single-knee samples. One primary reason is the imprecise fitting of straight lines onto curved segments. The fitted curve with the lowest RMSE is not a reliable indicator of the presence of a knee point. The \textit{S}-method not only shares the same limitations as the \textit{AL}-method, but it is also limited to working with simple concave curves. As a result, it consistently forecasts the initial point as a knee point. This explains its poor performance on the $ng$ test set (\textit{FT9}), which always has an elbow point before a knee point as shown in Fig. \ref{fig:sig_in_diff_x_intervals}. 

Regardless of whether rotation or projection is used, \textit{Kneedle} produces consistent results for all test sets. One must define a single sensitivity value to compute a unique threshold at every local maximum point of the transformed data curve. However, determining a universal sensitivity value that effectively applies to all local maximum points of a sample is challenging due to the variation in the transformed data magnitude. Furthermore, \textit{Kneedle} is highly susceptible to noise, which can erroneously classify a spike as local maximum and thus consider a noisy data point as a candidate. It is worth noting that the knee point may not always be reached at a local maximum. The \textit{Kneedle} algorithm has a tendency to detect multiple false negatives.

Finally, even in the harshest case of setting allowable index error equal to 1, our model reaches 0.55, 0.57, and 0.61 $F_{1}$ scores on the test sets, which are double the best results achieved by other methods. 

\section{Conclusion}
\label{sec:conclusion}

In this paper, we introduce a novel mathematical definition for a knee point in discrete data sets. We show and explain the necessity of rescaling the data. We develop a benchmark dataset that provides noisy data within the original data range, along with ground truth labels that are independent of any underlying algorithm/techniques. We believe that this benchmark dataset can serve as a common ground for evaluating future knee detection designs. We propose a new model, \textit{UNetConv}, for detecting knee points in discrete data sets, and compare its performance with existing approaches using the developed benchmark dataset. Our results indicate that \textit{UNetConv} outperforms other existing methods and exhibits exceptional performance on unseen data.

The limitations of this study include: (1) The noise introduced to the samples is Gaussian noise, which can make it relatively easier for the model to detect knee points due to its distinct characteristics. However, in real-world scenarios, the noise may not always follow a Gaussian distribution, which can affect the accuracy and generalizability of the model's performance. (2) There is a lack of versatility in the functions chosen to generate samples. As a result, there are numerous curve shapes that have yet to be explored by the model. (3) In all the synthetic data sets, the highest number of knee points observed in a curve is five. It remains uncertain whether the proposed model is capable of dealing with scenarios involving more than five knee points.

Considering the aforementioned limitations, it is essential to conduct an extensive
and in-depth investigation. Future work includes but is not limited to incorporating a
wider range of samples with varying levels of noise. This will provide valuable insights
into the model’s robustness and its ability to handle noisy data, thereby revealing the
model’s sensitivity to noise.


{\small
\bibliographystyle{ieee_fullname}
\bibliography{egbib}
}



\end{document}